%% file: main.tex
\documentclass[11pt]{article}

\usepackage{acl}
\usepackage{times}
\usepackage{latexsym}
\usepackage[T5]{fontenc}
\usepackage[utf8]{inputenc}
\usepackage{microtype}
\usepackage{inconsolata}
\usepackage{graphicx}
\usepackage{float}
\usepackage{subcaption}
\usepackage{tcolorbox}
\usepackage{amsmath}

\title{VLSP 2025 MLQA-TSR Challenge: Vietnamese Multimodal Legal Question Answering on Traffic Sign Regulation}

\author{
 \textbf{Son T. Luu\textsuperscript{1,2,3}},
 \textbf{Trung Vo\textsuperscript{1}},
 \textbf{Hiep Nguyen\textsuperscript{1}},
 \textbf{Khanh Quoc Tran\textsuperscript{2,3}},
\\
 \textbf{Kiet Van Nguyen\textsuperscript{2,3}},
 \textbf{Vu Tran\textsuperscript{1}},
 \textbf{Ngan Luu-Thuy Nguyen\textsuperscript{2,3}},
 \textbf{Le-Minh Nguyen\textsuperscript{1}}
\\
 \textsuperscript{1}Japan Advanced Institute of Science and Technology, Ishikawa, Japan, \\
 \textsuperscript{2}University of Information Technology, Ho Chi Minh City, Vietnam, \\
 \textsuperscript{3}Vietnam National University, Ho Chi Minh City, Vietnam
\\
 \small{
   \textbf{Correspondence:} sonlt@jaist.ac.jp, ngannlt@uit.edu.vn, nguyenml@jaist.ac.jp
 }
}

\begin{document}
\maketitle
\begin{abstract}
This paper presents the VLSP 2025 MLQA-TSR - the multimodal legal question answering on traffic sign regulation shared task at VLSP 2025. VLSP 2025 MLQA-TSR comprises two subtasks: multimodal legal retrieval and multimodal question answering. The goal is to advance research on Vietnamese multimodal legal text processing and to provide a benchmark dataset for building and evaluating intelligent systems in multimodal legal domains, with a focus on traffic sign regulation in Vietnam. The best-reported results on VLSP 2025 MLQA-TSR are an F2 score of 64.55\% for multimodal legal retrieval and an accuracy of 86.30\% for multimodal question answering.

\end{abstract}

\input{section/introduction.tex}
\input{section/task.tex}
\input{section/dataset.tex}
\input{section/method.tex}
\input{section/results.tex}
\input{section/conclusion.tex}

% \section*{Acknowledgments}
% This research is supported by DS2025-26-01

\bibliography{references}

% \appendix

% \section{Example Appendix}
% \label{sec:appendix}

% This is an appendix.

\end{document}

%% file: section/introduction.tex
\section{Introduction}
\label{intro}
Multimodal Question Answering (QA) is a challenging task in Natural Language Processing (NLP) that requires systems to understand and integrate multiple data types—such as text and images—to extract the correct information. Multimodal QA plays an important role in building intelligent systems because it offers a natural, user-friendly way for humans to search for information \cite{nguyen2023evjvqa}. Moreover, legal text processing is also difficult for NLP: legal language is highly formal, structurally complex, and rich in specialized terminology that presupposes substantial knowledge of legal concepts and principles \cite{10299488}. To answer legal questions correctly, systems must not only have an in-depth understanding of legal documents but also perform reasoning over legal information to arrive at the correct conclusions.

In recent years, many competitions have been organized on legal text processing to boost research in artificial intelligence for legal text processing. The COLIEE \cite{goebel2024overview} is an annual competition about legal text processing that targets the in-depth legal text understanding tasks like legal entailment and legal question answering in the English language. Similar to the COLIEE series, the ALQAC \cite{11063484} and VLSP-2023-LTER \cite{tran2024vlsp} are two shared tasks about legal text processing, including legal retrieval, legal question-answering, and legal textual entailment in Vietnamese legal documents. These competitions provide valuable benchmark datasets and a forum for research attempts in legal processing. However, these competitions focus on text only for legal domains, which motivates us to construct a multimodal competition about legal processing.  

\begin{figure*}[ht!]
    \centering
    \includegraphics[width=.8\textwidth]{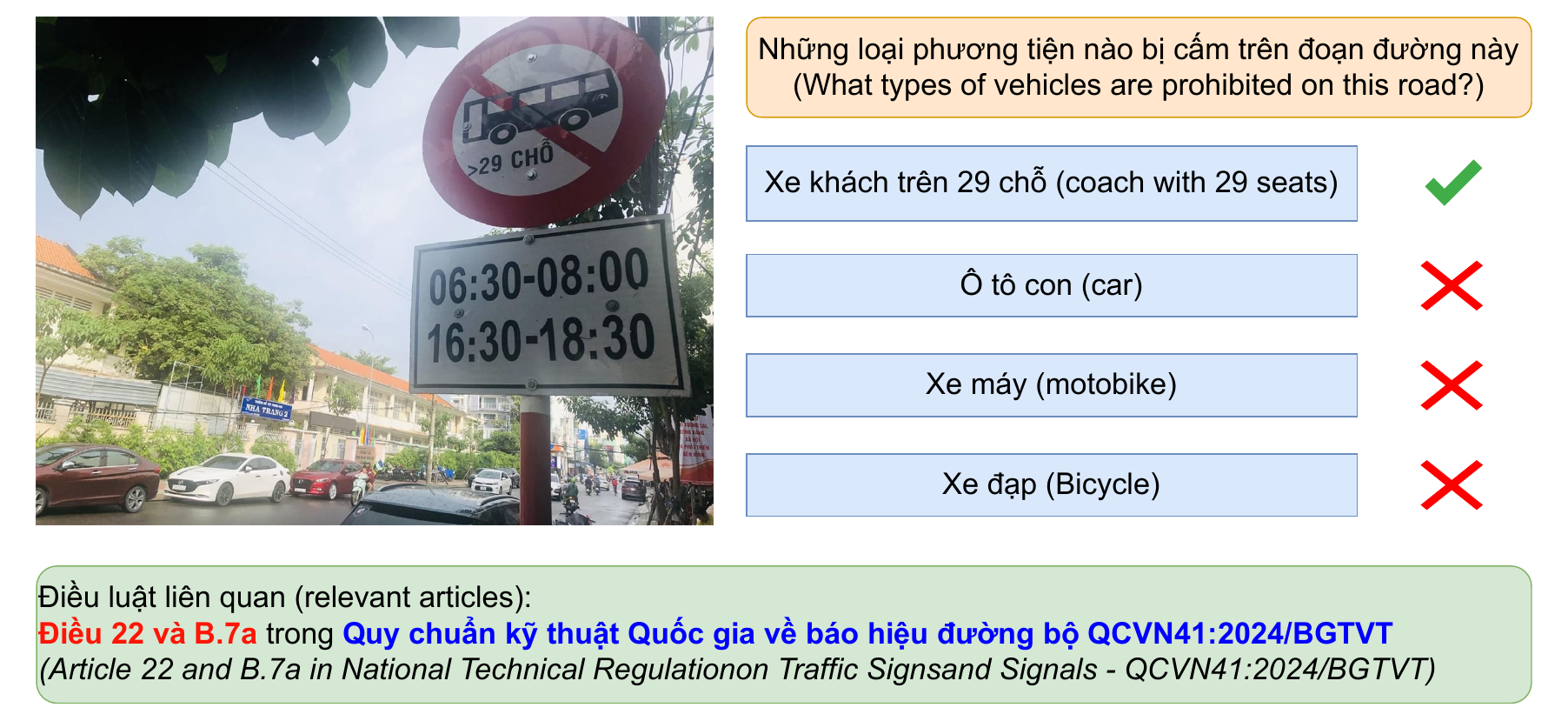}
    \caption{A sample of a legal question about a traffic sign.}
    \label{fig_sample}
\end{figure*}

Building on previous Vietnamese legal NLP competitions such as \textsc{ALQAC} \cite{11063484}, \textbf{VLSP 2025 MLQA-TSR} (VLSP 2025 \textbf{M}ultimodal \textbf{L}egal \textbf{Q}uestion \textbf{A}nswering on \textbf{T}raffic \textbf{S}ign \textbf{R}egulation) introduces a multimodal task centered on traffic sign regulation. As illustrated in Figure \ref{fig_sample}, answering a legal question requires understanding both the textual question and the accompanying image of traffic signs and then consulting the relevant statutes on road traffic and safety \cite{36/2024/QH15} as well as the \emph{Regulation on Traffic Signs and Signals} \cite{QCVN41:2024/BGTVT}. VLSP 2025 MLQA-TSR aims to spur the development of intelligent systems that can interpret multimodal legal inputs and retrieve correct answers to users' questions. The challenge comprises two subtasks: \emph{Legal Retrieval} (SubTask 1) and \emph{Legal Question Answering} (SubTask 2), and is hosted on the CodaBench platform \cite{codabench}.

Overall, this paper provides an overview of the VLSP 2025 MLQA-TSR share task. We summarize two main contributions in this paper as: 
\begin{itemize}
    \item First, we provide a benchmark dataset about multimodal legal question answering on traffic sign regulation in Vietnam. The dataset consists of two multimodal tasks: legal retrieval and legal question answering. 
    \item Second, we organize the shared task VLSP 2025 MLQA-TSR at VLSP 2025. We obtained 13 team submissions for SubTask 1 and 19 team submissions for SubTask 2. 
\end{itemize}

The remainder of the paper is organized as follows. Section \ref{task} describes the two subtasks in detail. Section \ref{dataset} presents the data construction process and an overall analysis. Section \ref{method} summarizes participant approaches and the baseline used in the competition. Section \ref{results} reports the results and rankings. Finally, Section \ref{conclusion} concludes the paper and outlines future directions.

%% file: section/task.tex
\section{Task description}
\label{task}
VLSP 2025 MLQA-TSR shared task focuses on enhancing the ability of computers to understand legal text in multimodal scenarios about traffic sign regulation. The shared task includes two subtasks: multimodal legal retrieval (Subtask 1) and multimodal legal question answering (Subtask 2). 

\subsection{Subtask 1: Multimodal Legal Retrieval}
\paragraph{Task definition:} Given a multimodal question $q=(q_{\text{text}}, q_{\text{image}})$ consisting of two components, where $q_{\text{text}}$ is the question in textual form and $q_{\text{image}}$ is the query image corresponding to $q_{\text{text}}$—and a law database $\mathcal{D}=\{d_i\mid i=1,\dots,n\}$ containing articles from legal documents, the goal is to determine a ranked list of relevant articles $\mathcal{R}=\{d_i\mid d_i\in\mathcal{D},\, i=1,\dots,k,\, k\le n\}$ for the question $q$. Each article $d_i$ may be text-only or multimodal (e.g., text, images, and/or tables).

\paragraph{Evaluation metric:} We use the F2-score for assessing the performance of the retrieval model since the F2-score places more in recall, which is concerned about false negatives more than false positives. The F2-score for a question q is computed as Equation \ref{eq_f2}. The final F2-score is determined by averaging the F2-score over the evaluation sets. 

\begin{equation}
    \label{eq_precision}
    Precision_{q}=\frac{\#correct\_retrieved\_articles}{\#total\_retrieved\_articles}
\end{equation}

\begin{equation}
    \label{eq_recall}
    Recall_{q}=\frac{\#correct\_retrieved\_articles}{\#total\_relevant\_articles}
\end{equation}

\begin{equation}
    \label{eq_f2}
    F2_{q}=5*\frac{Precision_{q}*Recall_{q}}{4*Precision_{q}+Recall_{q}}
\end{equation}

\subsection{Subtask 2: Multimodal Legal Question Answering}
\paragraph{Task definition:} Given a multimodal question $q=(q_{\text{text}}, q_{\text{image}})$—with $q_{\text{text}}$ as the textual query and $q_{\text{image}}$ as the corresponding image—and a list of relevant articles $\mathcal{R}=\{d_i\mid d_i\in\mathcal{D},\, i=1,\dots,k,\, k\le n\}$ for $q$, the objective is to predict the correct answer $a$ from four multiple-choice options for the multimodal question $q$.

\paragraph{Evaluation metric:} Subtask 2 is formulated as multiple-choice question answering. Therefore, we employ \emph{Accuracy} as the primary metric, measuring the proportion of correctly predicted answers over the evaluation set. The accuracy is determined as Equation \ref{eq_acc}

\begin{equation}
    \label{eq_acc}
    Accuracy=\frac{\#total\_correct\_answers}{\#total\_questions}
\end{equation}

%% file: section/dataset.tex
\section{Dataset}
\label{dataset}

\subsection{Data Construction}
\begin{figure*}[ht!]
    \centering
    \includegraphics[width=\textwidth]{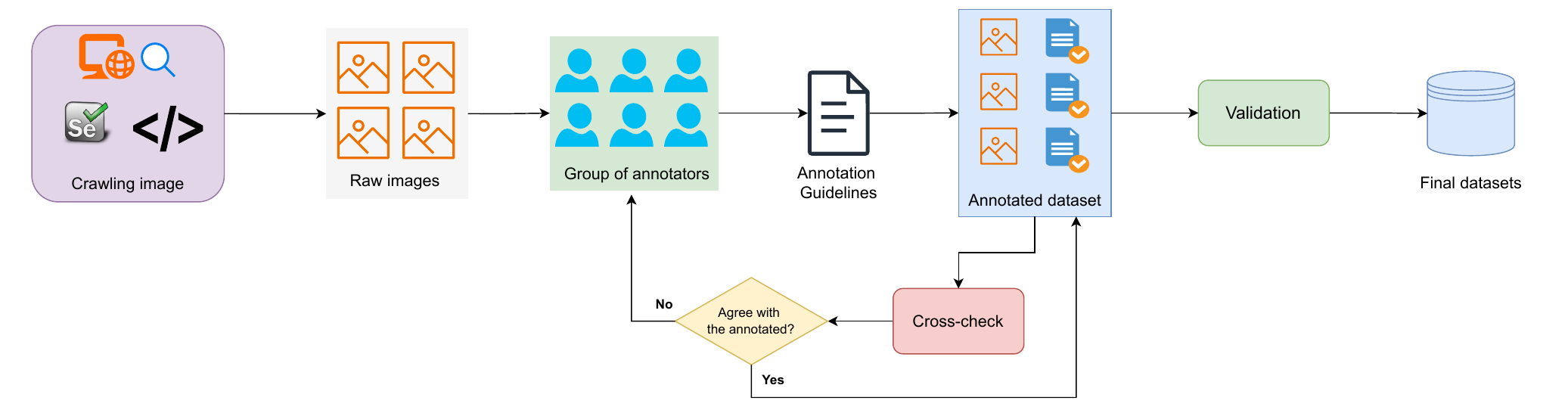}
    \caption{Data Creation Process.}
    \label{fig_data_creation}
\end{figure*}

Figure \ref{fig_data_creation} shows the overview of the data construction process for the VLSP 2025 MLQA-TSR share task. The process consists of three main stages as described follows:

\begin{itemize}
    \item \textbf{Stage 1 - Data Collection:} We search the images of traffic signs on streets in Vietnam on Google search, then we collect them automatically by using the Selenium tool based on HTML processing. After stage 1, we have a raw image set that contains various traffic signs on the street, and also non-relevant images to traffic signs. We then manually remove the images that are not relevant to traffic signs. 
    \item \textbf{Stage 2 - Data Annotation:} We hire a group of 8 annotators who are undergraduate students and give them the annotation guidelines. After reading the annotation guidelines, we let the annotator create a question and an answer based on the image of traffic signs. First, annotators are required to read the Regulation on Traffic Signs and Signals and the Road Traffic and Safety Law, then give a list of the most relevant articles to the question. Second, the annotators make the correct answer according to the question. There are two kinds of questions: multiple-choice and yes/no questions. For the multiple-choice, annotators are required to create four different choices and mark the correct one. The choices of Yes/No questions have only two options, including "Đúng" indicating "Yes" and "Sai" representing "No".
    \item \textbf{Stage 3 - Cross-checking:} We let the annotators perform cross-checking among the annotated data. One will check the correctness of the answers and relevant articles to the question and traffic sign images, syntax, and typos of the questions and answers of others. If the annotator disagrees with an annotated sample, the disagreement sample will be sent back to the group of annotators for re-annotating. 
    \item \textbf{Stage 4 - Validation:} We let the annotators perform final validation of annotated data. The main criteria for validation include: the consistency between the question and the traffic signs, the correctness of the relevant articles and the answers, and the typos and syntax in the question and answers. If any samples do not satisfy the criteria, we remove them from the dataset. 
\end{itemize}

The final dataset is split into three sets: training, public test, and private test sets to serve for the shared task. Besides, we also provide the law database, including the articles in both the National Regulation on Traffic Signs and Signals (\textbf{QCVN 41:2024/BGTVT}) and the Law on Road Traffic Order and Safety (\textbf{36/2024/QH15}). In the National Regulation on Traffic Signs and Signals, we represent the image in the articles with a format \textit{<<IMAGE: image\_file.jpg /IMAGE>>} and the table as the following format \textit{<<TABLE: table\_html\_code /TABLE>>}. The law database is provided to the participants as a JSON file format along with a directory containing corresponding images. 

\subsection{Data Analysis}
Table \ref{tbl_law_db_info} summarizes the overall information about the two legal documents used in the law database. It can be seen that the document \textbf{QCVN 41:2024/BGTVT} - "National Regulation on Traffic Signs and Signals" contains both image and table data in the article, while the \textbf{36/2024/QH15} - "Law on Road Traffic Order and Safety" only has text in the document. Also, the number of articles in the National Regulation on Traffic Signs and Signals is significantly more than the Law on Road Traffic Order and Safety, since the National Regulation on Traffic Signs and Signals contains a detailed description of the technical specifications and the meaning of various traffic signs in road traffic in Vietnam. 

Next, Table \ref{tbl_data_stat} illustrates the overall statistics about the three sets used in the VLSP 2025 - MLQA-TSR. In the training and public test sets, the proportion between multiple-choice and Yes/No questions is 6/3, while this proportion is 5-5 in the private test to ensure the objective performance of the question-answering models for the type of question. Also, the average length of a question in the private test set is higher than in the training and public test sets, challenging the generability of question answering models in generating correct answers for the questions (The length of the question is computed according to the number of tokens in the question. We segment the question text into token-level by using the Pyvi\footnote{\url{https://pypi.org/project/pyvi/}}).

\begin{table}[H]
\centering
    \caption{Overview information about two legal documents in the law database}
    \label{tbl_law_db_info}
    \resizebox{\columnwidth}{!}{
        \begin{tabular}{lrr}
        \hline
        \textbf{Law ID} & \multicolumn{1}{l}{\textbf{QCVN 41:2024/BGTVT}}                                                                  & \multicolumn{1}{l}{\textbf{36/2024/QH15}}                                                            \\ \hline
        Law Name        & \multicolumn{1}{l}{\begin{tabular}[c]{@{}l@{}}National Regulation \\ on Traffic Signs and Signals.\end{tabular}} & \multicolumn{1}{l}{\begin{tabular}[c]{@{}l@{}}Law on Road Traffic \\ Order and Safety.\end{tabular}} \\
        \# Articles     & 313                                                                                                              & 89                                                                                                   \\
        \# Image        & 761                                                                                                              & 0                                                                                                    \\
        \# Table        & 212                                                                                                              & 0                                                                                                    \\ \hline
        \end{tabular}
    }
\end{table}

\begin{figure*}[ht!]
    \begin{subfigure}{0.33\textwidth}
        \includegraphics[width=\textwidth]{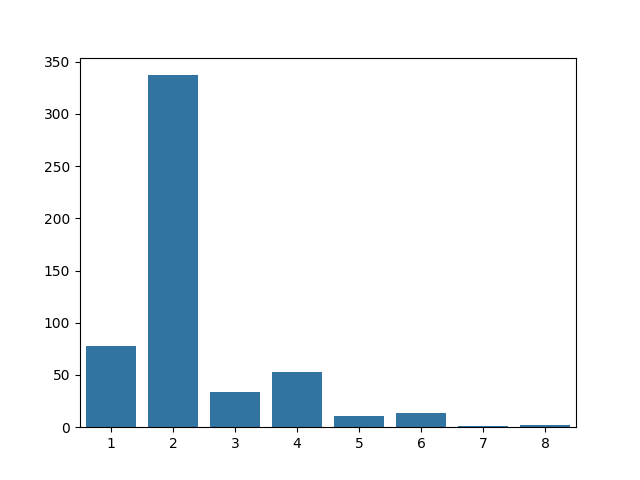}
        \caption{Training set}
    \end{subfigure}
    \begin{subfigure}{0.33\textwidth}
        \includegraphics[width=\textwidth]{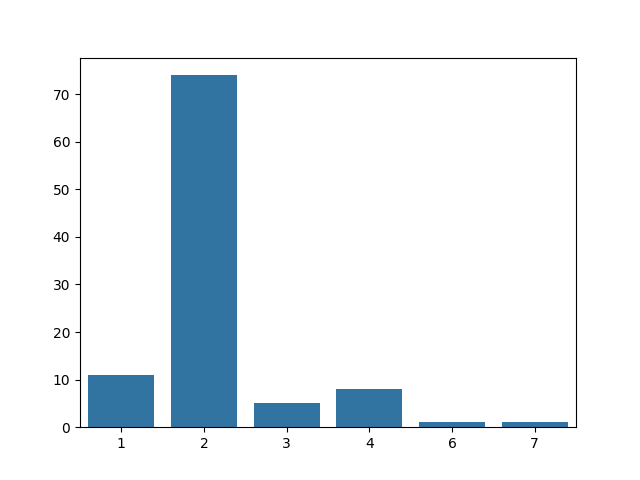}
        \caption{Public Test}
    \end{subfigure}
    \begin{subfigure}{0.33\textwidth}
        \includegraphics[width=\textwidth]{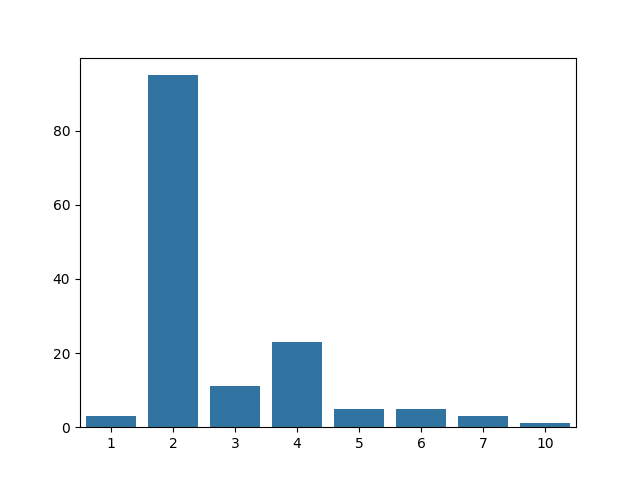}
        \caption{Private Test}
    \end{subfigure}\par\medskip
    \caption{Distribution of relevant articles in three sets}
    \label{fig_article_dist}
\end{figure*}

\begin{table}[H]
    \centering
    \caption{Statistics about the provided dataset in VLSP 2025 MLQA-TSR}
    \label{tbl_data_stat}
    \resizebox{\columnwidth}{!}{
    \begin{tabular}{lrrr}
    \hline
                               & \multicolumn{1}{l}{\textbf{Train}} & \multicolumn{1}{l}{\textbf{Public Test}} & \multicolumn{1}{l}{\textbf{Private Test}} \\ \hline
    \# Total questions                 & 530                                & 100                                      & 146                                       \\
    \# Total images                 & 304                                & 90                                      & 104                                       \\
    \# Multiple choices questions       & 376                                & 65                                       & 74                                        \\
    \# Yes/No questions         & 154                                & 35                                       & 72                                        \\ \hline
    Max length of question     & 69                                 & 42                                       & 76                                        \\
    Min length of question     & 5                                  & 5                                        & 8                                         \\
    Average length of question & 16.95                              & 14.96                                    & 27.00                                     \\
    Max relevant articles      & 8                                  & 7                                        & 10                                        \\
    Min relevant articles      & 1                                  & 1                                        & 1                                         \\
    Average relevant articles  & 2.31                               & 2.19                                     & 2.76                                      \\ \hline
    \end{tabular}
    }
\end{table}

On the other hand, the number of relevant articles per question on the three sets is similar, with around two articles for a question. As shown in Figure \ref{fig_article_dist}, most questions in the datasets have two relevant articles. For the training and public test sets, the number of relevant articles usually falls into 1 to 2 articles, while it often falls into about 2 to 4 articles in the private test. Overall, the distribution of data in the private test is slightly different from the training and public test sets to ensure the objective of the model and avoid overfitting.

%% file: section/method.tex
\section{Method}
\label{method}
\subsection{Baseline Methods} 
We adopt \textbf{BGE Visualized} \cite{zhou-etal-2024-vista} as a strong and efficient baseline for multimodal retrieval.
We encode each article in the law database into a \(d\)-dimensional embedding (we use \(d{=}1024\)), yielding a matrix of shape \((K, d)\), where \(K\) is the number of articles.
Given a query consisting of the textual question and the associated image, we encode it into a single \(d\)-dimensional embedding and compute dot-product similarities against all article embeddings.
We then retrieve the \emph{top-5} most similar articles as candidates for the query. 

For QA, we use \textbf{Vintern} \cite{doan2024vintern1befficientmultimodallarge}, a Vietnamese Multimodal Large Language Model (MLLM).
We prompt the \texttt{Vintern-3B-beta} checkpoint to generate answers.
For the two question types (multiple choice and Yes/No), we employ the following prompt templates:

\begin{tcolorbox}[boxrule=0pt, width=\linewidth, title=Multiple-choice questions]
% \footnotesize{
<image> \\
<question> \\
A. Lựa chọn 1. (\textit{English}: Choice 1) \\
B. Lựa chọn 2. (\textit{English}: Choice 2) \\
C. Lựa chọn 3. (\textit{English}: Choice 3) \\
D. Lựa chọn 4. (\textit{English}: Choice 4) \\

\textit{Trả lời bằng một trong bốn đáp án: A, B, C hoặc D. Không giải thích thêm} \\
(\textit{English}: Answer by one of four choices: A, B, C, or D without explanation)
% }

\end{tcolorbox}

\begin{tcolorbox}[boxrule=0pt, width=\linewidth, title=Yes/No questions]
% \footnotesize{
<image> \\
<question> \\

\textit{Trả lời bằng một trong hai đáp án: Đúng hoặc Sai. Không giải thích thêm.} \\
(\textit{English}: Answer by one of two values: Yes or No without explanation)
% }

\end{tcolorbox}

\subsection{Proposed Methods by Participants} 
This section summarizes the methodologies proposed by the top--5 teams for each subtask in VLSP~2025 MLQA-TSR.
Participants were required to use only open-source LLMs whose code and model weights are publicly available (e.g., GitHub, Hugging Face); commercial LLMs (e.g., ChatGPT, Gemini, Claude) were not allowed.
Additionally, the use of any external data beyond the competition resources was prohibited.
Below are the reported approaches:
\begin{itemize}
    \item \textbf{SmartbotIC:}
    The team uses \texttt{Qwen2.5-VL} and \texttt{InternVL3} with zero-shot prompting to generate answers from the input question and image, together with the retrieved articles. To improve efficiency and quality, they apply several preprocessing steps, including concatenating multiple images into a single image and converting HTML tables to Markdown to reduce input length.
    
    \item \textbf{Berry:}
    For retrieval, the team encodes questions and choices with \texttt{Jina Embeddings}  (text) and uses \texttt{C-RADIOv2-B}  to encode images, while OWLv2 detects traffic-sign objects whose features are also embedded. All vectors are stored in a Qdrant vector database; subtask~1 results are obtained via vector search to return the top-\(k\) most similar candidates. For QA, they apply few-shot prompting with examples retrieved from the vector store and generate final answers using the \texttt{Llama~4 Maverick}.
    
    \item \textbf{Tanka\_CDS:}
    The team preprocesses the law database by converting HTML tables to normalized text, chunking, and embedding articles with CLIP. For retrieval, \texttt{YOLOv8n} is fine-tuned to extract key regions in traffic-sign images as patches; \texttt{LLaMA3.2-Vision} then encodes the question with the corresponding patches. For QA, \texttt{LLaMA3.2-Vision} generates answers based on the question, image, and the retrieved articles.
    
    \item \textbf{chmod+x:}
    The team models a \emph{heterogeneous graph} to represent relationships among images, text, and tables in the law database. They then apply the \texttt{Jina Reranker} to re-rank visual documents and perform graph matching combined with similarity search to retrieve the top-\(k\) candidates for a given question and image.
    They further propose a dynamic top-\(k\) filtering mechanism based on counting relevant traffic signs to adapt the candidate set size to question complexity.
    For QA, they use an ensemble of two multimodal LLMs: Qwen2.5-VL-7B and InternVL3-8B.
    
    \item \textbf{Metamorphic:}
    The team designs a systematic workflow to produce final answers from user inputs (question, image, choices).
    They construct graph representations for images (\emph{ImageSubGraph}) and for articles (\emph{ArticleSubGraph}) to retrieve enriched information (e.g., scene descriptions and fine-grained traffic-law content). This information is merged into a unified prompt to guide the LLM. In the framework, \texttt{Gemma2-27B} and \texttt{DeepSeek} are used to encode and process multimodal inputs for QA. \texttt{YOLO} is also employed for traffic-sign detection as an additional feature.
    
    \item \textbf{LifeIsTough:}
    The team preprocesses the law database by converting HTML tables to Markdown and normalizing text.
    For database images, \texttt{Gemma-3-12B} is used to crop regions so that only traffic signs are retained.
    Text and images are then encoded with \texttt{CLIP} and stored in Qdrant. For retrieval, the team first applies \texttt{YOLOE} to detect traffic signs in the input image, then validates the detections against the input question using \texttt{Gemma-3-12B}. The enriched query is encoded with \texttt{CLIP} to construct a query vector, which is used to search the law database (already embedded) for relevant articles. For QA, the retrieved knowledge is combined with the question in the prompt, and \texttt{Gemma-3-12B} generates the final answer.
    
    \item \textbf{TechNova:}
    For retrieval, the team implements a two-branch architecture. The first branch generates separate image and text embeddings for the entire training set.
    The second branch builds a dense corpus retriever by indexing text-chunk embeddings from the full law database in a FAISS index. To retrieve relevant articles, the query (image + question) is fused into a multimodal embedding and compared against article embeddings. The team uses \texttt{gme-Qwen2-VL-2B-Instruct} to produce retrieval embeddings.
    For QA, they employ \texttt{Qwen2.5-VL-72B-Instruct} with Chain-of-Thought prompting: first describe the visual scene, then apply the legal context, and finally reason to a conclusion.
\end{itemize}

Overall, the Multimodal Legal Retrieval Task (Subtask 1) relies on constructing a contextual multimodal embedding space to perform similarity search or re-ranking. Robust multimodal embedding models such as CLIP \cite{radford2021learningtransferablevisualmodels}, Jina Embedding \cite{günther2025jinaembeddingsv4universalembeddingsmultimodal}, C-RADIOv2 \cite{heinrich2025radiov2} and multimodal LLMs like Gemma-3 \cite{gemmateam2025gemma3technicalreport} and Qwen2-VL \cite{wang2024qwen2vlenhancingvisionlanguagemodels}. To enhance the accuracy of the retrieval task, participants often employ traffic sign detection models like OWL \cite{minderer2023scaling} or YOLO \cite{yaseen2024yolov8indepthexplorationinternal} to filter the key information about traffic signs in the images or crop them by Gemma-3 \cite{gemmateam2025gemma3technicalreport}. In addition, data preprocessing steps such as text normalization, image filtering, combination, concatenation, and HTML table transformation are also frequently used to enhance the performance of the model in vector embedding. In addition, the vector databases, such as Qdrant \footnote{\url{https://qdrant.tech/}} or FAISS \footnote{\url{https://github.com/facebookresearch/faiss}}, are usually used in a retrieval task to serve for vector-space similarity searching. Moreover, several teams use retrieval and searching on graphs to improve the performance of the retrieval task by efficiently representing the multimodal data in graphs to capture semantic relation information. For the Multimodal Legal Question Answering Task (Subtask 2), the vision LLMs like Qwen2.5-VL \cite{bai2025qwen25vltechnicalreport}, InternVL3 \cite{zhu2025internvl3exploringadvancedtraining}, Gemma2 \cite{gemmateam2024gemma2improvingopen}, and LLaMa3.2-Vision \cite{grattafiori2024llama3herdmodels} are used by almost all participants, indicating the robustness and efficiency of vision LLMs for multimodal QA. Zero-shot prompting, Chain-of-Thought \cite{wei2022chain}, and few-shot prompting are mostly used techniques by participants to instruct vision LLMs in generating the answer. 

%% file: section/results.tex
\section{Results}
\label{results}
Table \ref{tbl_result_subtask1} shows the ranking results of participants for Subtask 1 on the private test. The top 1 team - \textbf{LifeIsTough}, with the efficient retrieval method that filters the key features in the traffic sign image via cropping by LLM and latent vector embedding construction by CLIP \cite{radford2021learningtransferablevisualmodels}, achieves the highest results with 64.55\% by F2 score. The \textbf{chmod+x} team is runner-up with 61.13\%, and \textbf{TechNoVa} places third with 59.91\% by F2 score. Additionally, there is a significant gap between the top 5 teams with others in Subtask 1, where the top 5 teams attain a performance by F2 score of more than 50\%, and others obtain lower than 50\% of performance by F2 score. All teams in Subtask 1 obtain performance better than the baseline methods, indicating the efficiency of the proposed method for the legal retrieval task. Since the highest results for the legal retrieval task are approximately 65\%, there is still room for further improvement in this task. 

Next, Table \ref{tbl_result_subtask2} illustrates the ranking of participants for Subtask 2 on the private test\footnote{We report the results that performance over the baseline}. The top 1 team - \textbf{Berry} obtains optimistic results for this task with 86.30\% by Accuracy by employing the LLama4-Maverick with efficient few-shot prompting. \textbf{SmartbotIC} is the runner-up team with 83.56\%, and \textbf{TechNova} obtains $3^{rd}$ rank with 78.08\% by Accuracy. Overall, it can be seen that the gap between the top 5 teams with others in Subtask 2 is not as much as in Subtask 1, indicating the efficiency and robustness of the proposed methodologies by participants for the multimodal legal question answering. 

\begin{table}[H]
    \centering
    \caption{Results for subtask 1 - Multimodal Legal Retrieval}
    \label{tbl_result_subtask1}
    \begin{tabular}{lrr}
    \hline
    \textbf{Team name}   & \multicolumn{1}{l}{\textbf{F2 score}} & \multicolumn{1}{l}{\textbf{Rank}} \\ \hline
    \textbf{LifeIsTough} & \textbf{0.6455361395}                       & \textbf{1}                        \\
    \textbf{chmod+x}     & \textbf{0.6113748745}                       & \textbf{2}                        \\
    \textbf{TechNova}    & \textbf{0.5991697165}                       & \textbf{3}                        \\
    \textbf{SmartbotIC}  & \textbf{0.5790135683}                       & \textbf{4}                        \\
    \textbf{Berry}       & \textbf{0.5432150682}                       & \textbf{5}                        \\
    DHDD                 & 0.4511688070                                & 6                                 \\
    Tanka\_CDS           & 0.2459355607                                & 7                                 \\
    AIO\_VNM             & 0.2384666964                                & 8                                 \\
    MealsRetrieval       & 0.1548250424                                & 9                                 \\
    OpenCubee            & 0.1533444945                                & 10                                 \\
    Come4Win             & 0.1413907401                                & 11                                \\
    LexTraffic           & 0.1358808937                                & 12                                \\
    BASELINE             & 0.1276493485                                & 13                                \\ \hline
    \end{tabular}
\end{table}

\begin{table}[h]
    \centering
    \caption{Results for subtask 2 - Multimodal Legal Question Answering}
    \label{tbl_result_subtask2}
    \begin{tabular}{llr}
    \hline
    \textbf{Team name}   & \textbf{Accuracy score} & \multicolumn{1}{l}{\textbf{Rank}} \\ \hline
    \textbf{Berry}       & \textbf{0.8630136986}   & \textbf{1}                        \\
    \textbf{SmartbotIC}  & \textbf{0.8356164384}   & \textbf{2}                        \\
    \textbf{TechNova}    & \textbf{0.7808219178}   & \textbf{3}                        \\
    \textbf{Tanka\_CDS}  & \textbf{0.7328767123}   & \textbf{4}                        \\
    \textbf{Metamorphic} & \textbf{0.7260273973}   & \textbf{5}                        \\
    Hallucinators        & 0.7123287671            & 6                                 \\
    LifeIsTough          & 0.6712328767            & 7                                 \\
    chmod+x              & 0.6232876712            & 8                                 \\
    OpenCubee            & 0.6095890411            & 9                                 \\
    LexTraffic           & 0.5958904110            & 10                                \\
    AIO\_VNM             & 0.5684931507            & 11                                \\
    NaN                  & 0.5616438356            & 12                                \\
    SoftMind\_AIO        & 0.5000000000            & 13                                \\
    BASELINE             & 0.4520547945            & 14                                \\
    % Warriors             & 0.4383561644            & 15                                \\
    % N4                   & 0.4109589041            & 16                                \\
    % DHDD                 & 0.3561643836            & 17                                \\
    % MealsRetrieval       & 0.2191780822            & 18                                \\
    % UIT\_NLP\_Group10    & 0.1232876712            & 19                                \\ \hline
    \hline
    \end{tabular}
\end{table}

In comparison with ALQAC 2024 \cite{11063484} in Vietnamese language and COLIEE 2024 \cite{goebel2024overview} in English language - the two competitions about legal document processing, it can be seen that the best performance on the legal retrieval task is about 87\% by F2 score at ALQAC 2024, and 44\% by F2 score at COLIEE 2024 (Task 1), while the performance of legal question answering task is significantly higher with 98\% by Accuracy at ALQAC 2024 and 82\% by Accuracy at COLIEE 2024 (Task 4). In general, the legal retrieval task is more challenging than question answering, since legal documents have a complex structure and specialized legal terminologies that require an in-depth understanding between the users' queries and the legal documents to extract correct and valuable information. In the scenario of multimodal, the retrieval system not only focuses on text but is also concerned about the latent information from the image to provide the correct answer. 

%% file: section/conclusion.tex
\section{Conclusion}
\label{conclusion}
This paper introduces VLSP 2025—MLQA-TSR, a new multimodal shared task in legal text processing designed to advance research on low-resource languages, with a primary focus on Vietnamese. The task comprises two subtasks: (1) multimodal legal retrieval and (2) multimodal question answering. The best-performing systems achieve an F2 score of 64.55\% on the multimodal legal retrieval subtask and an accuracy of 86.30\% on the multimodal question answering subtask. VLSP 2025—MLQA-TSR attracted a range of innovative methodologies leveraging state-of-the-art models across both subtasks, offering valuable momentum for research in Vietnamese multimodal legal text processing. Finally, VLSP 2025—MLQA-TSR provides a benchmark dataset for building and evaluating intelligent systems in the legal domain and multimodal tasks in Vietnamese, specifically centered on traffic sign regulation. The dataset and baseline code are published at \url{https://github.com/sonlam1102/VLSP2025-MLQA-TSR}.